\documentclass[runningheads]{llncs}
\usepackage{CJKutf8}
\usepackage{graphicx}
\begin{document}
\title{LSTM Based Sentiment Analysis for Cryptocurrency Prediction}
%
%

\author{Xin Huang\inst{1}, Wenbin Zhang\inst{1}, Xuejiao Tang\inst{2}, Mingli Zhang\inst{3}, \\Jayachander Surbiryala\inst{4}, Vasileios Iosifidis\inst{2}, Zhen Liu\inst{5}\and Ji Zhang\inst{6}}

\authorrunning{X. Huang, W. Zhang, X. Tang, M. Zhang, J. Surbiryala et al.}

\institute{University of Maryland, Baltimore County, USA \\
\and
Leibniz University Hannover, Germany\\
\and
McGill University, Canada $^4$University of Stavanger, Norway\\
$^5$Guangdong Pharmaceutical University, China\\
$^6$University of Southern Queensland, Australia\\
\email{$^1$\{xinh1,wenbinzhang\}@umbc.edu,\\$^2$\{xuejiao.tang,iosifidis\}@stud.uni-hannover.de,\\$^3$mingli.zhang@mcgill.ca, $^4$jayachander.surbiryala@uis.no, $^5$liu.zhen@gdpu.edu.cn, $^6$ji.zhang@usq.edu.au}
}
\maketitle              

\vspace{-0.4cm} 
\begin{abstract}
Recent studies in big data analytics and natural language processing develop automatic techniques in analyzing sentiment in the social media information. In addition, the growing user base of social media and the high volume of posts also provide valuable sentiment information to predict the price fluctuation of the cryptocurrency. This research is directed to predicting the volatile price movement of cryptocurrency by analyzing the sentiment in social media and finding the correlation between them. While previous work has been developed to analyze sentiment in English social media posts, we propose a method to identify the sentiment of the Chinese social media posts from the most popular Chinese social media platform Sina-Weibo. We develop the pipeline to capture Weibo posts, describe the creation of the crypto-specific sentiment dictionary, and propose a long short-term memory (LSTM) based recurrent neural network along with the historical cryptocurrency price movement to predict the price trend for future time frames. The conducted experiments demonstrate the proposed approach outperforms the state of the art auto regressive based model by 18.5\% in precision and 15.4\% in recall.
\end{abstract}
%
%
%
\vspace{-0.85cm}
\section{Introduction}
\vspace{-0.2cm}
Since the birth of Bitcoin, there has been an enormous rise and interest in the cryptocurrency, a decentralized digital asset developed by the blockchain technology. This digital currency draws a lot of attention due to its volatility which provides the opportunity for digital trading with high return. The total market capitalization of cryptocurrencies has increased from 1 billion dollars to 400 billion dollars in the past decade, with the number still increasing.  

On the other hand, the emergence of social media such as Twitter, Reddit and Facebook also makes the latest news and social media posts about financial markets widely accessible. Investors have therefore been utilizing such a variety of digital resources to make trading decisions. Previous studies discovered evidence of such correlation between stock price movement and social media \cite{bollen2011}. Sentiment on cryptocurrency social media content with negative emotions, e.g., fear and sadness, neutral emotions, e.g., calm and not sure, or positive emotions, e.g., trust and happiness, can be used to predict cryptocurrency price fluctuations and further to assist the investment decision making. This paper focuses on this trending theme, proposing a recurrent neural network with long short-term memory (LSTM) by utilizing the sentiment analysis of social media to predict the real time price movement of the digital currency. 

\section{Related Work}
\vspace{-0.2cm}
Over the past decades, a variety of machine learning techniques have been developed to predict the price movement for the stock market using social media, such as opinion analysis of twitter feeds \cite{chen2019}. \cite{bollen2011} used neural networks and daily Twitter feeds as extra predictors to forecast the daily up and down changes in the closing values of the Dow Jones Industrial Average. Moreover, \cite{Pimprikar2017} found the Long Short-Term Memory (LSTM) combined with a Twitter sentiment analysis outperforms other machine learning models such as Support Vector Machine in predicting the stock price. 

Recent studies have also successfully applied sentiment analysis in various applications, such as predicting the movie revenues \cite{joshi2010}, analysing sentiment towards US presidential candidates in 2012 \cite{wang2012}. In the English sentiment analysis, Valence Aware Dictionary and sEntiment Reasoner (VADER) \cite{hutto2014} is used to classify sentiment in tweets. In this paper, we designed a crypto sentiment dictionary that is customized to cryptocurrency and Chinese Weibo posts. We use LSTM \cite{hochreiter1996} as the neural network learning layer and combine it with the sentiment analysis method to develop a crypto sentiment analyzer that can predict the price movement of cryptocurrency.

\vspace{-0.1cm}
\section{Methodology}
\begin{figure}[!htb]
    \vspace{-0.1cm}
    \centering
    \includegraphics[width=1.0\textwidth]{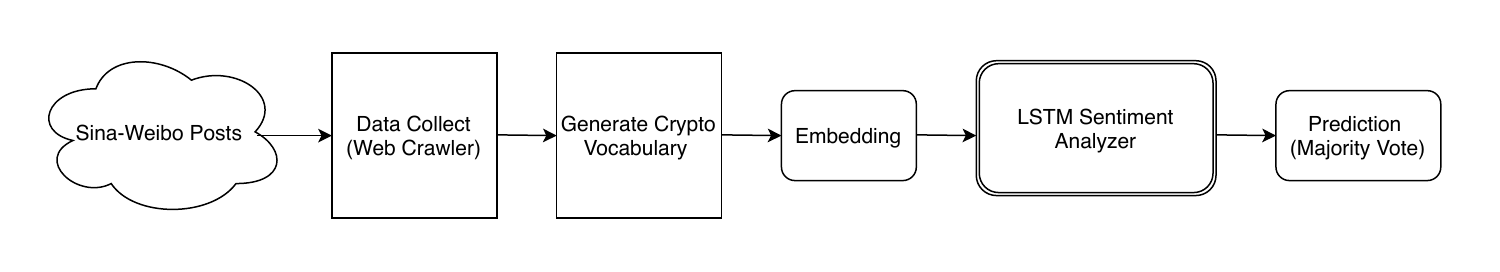}
    \caption{Architecture of LSTM based cryptocurrency sentiment analysis and price movement prediction. Crypto sentiment dictionary is created to generate the crypto word embedding, LSTM is to learn the sentiment information, and the majority voting on the output of LSTM sentiment analyzer is used to predict the price going up or down.}
    \label{fig:framework}
    \vspace{-0.5cm}
\end{figure}

\vspace{-0.1cm}
Figure \ref{fig:framework} shows the end to end architecture of the LSTM based sentiment deep learning model in predicting the real time price fluctuations. We crawled user posts from China's most popular social media platforms Sina-Weibo, and created a crypto-specific sentiment dictionary with domain-expert knowledge, and then the LSTM recurrent neural network was used to model the sentiment information and make real time prediction for the price trend.

\vspace{-0.2cm}
\subsection{Data Collection}
\vspace{-0.2cm}
Chinese investors exchange crypto information via news articles and social media platforms, especially using Sina-Weibo, Wechat and QQ groups. We collect a large-scale Weibo corpus from crawling Chinese microblogs on Sina-Weibo with the cryptocurrency keyword, in particular, Bitcoin, ETH or XPR. The number of crawled cryptocurrency tweets from Weibo is 24,000, as well as 70,000 comments to them, from the most recent 8 days. 

\subsection{Crypto Sentiment Dictionary}
The general sentiment dictionary created by natural language processing (NLP) is not applicable in the crypto domain. We introduce a novel way to build a crypto specific sentiment dictionary that can capture the unique  characteristics of the crypto social communities. Table \ref{tab:dict} shows an example dictionary that is particular to the Chinese crypto words in Sina-Weibo. 
The first step of generating a crypto sentiment dictionary is to create the vocabulary of the crypto words. We manually label the crawled Weibo posts with ranking, and use seed sentiment words selected by crypto domain experts to do bootstrapping, which adds high frequency new words in highly positive/negative weibo posts into the crypto corpus. 

After the crypto corpus is generated, we then create an index mapping dictionary in such a way that the frequently occurring crypto words are assigned lower indexes, similar to the traditional natural language processing. Finally we generate a crypto word encoding for each individual post and use that encoding vector as the training data for the RNN model in sentiment analysis.

\begin{CJK*}{UTF8}{gbsn}
\begin{table}[]
\vspace{-0.5cm}
\caption{Crypto specific sentiment dictionary building for Chinese words.}
\centering
\begin{tabular}{|l|l|l|}
\hline
Chinese & \textbf{Informal Translation} & \textbf{Implied Sentiment} \\
\hline
韭菜 & Bag Holder & Investors have profit loss \\
\hline
新高 & New High  & Investors are excited \\
\hline
下车 & Abandon Ship  & Investors rush to sell off \\
\hline
\end{tabular}
\label{tab:dict}
\end{table}
\end{CJK*}

\vspace{-0.8cm}
\subsection{LSTM based Sentiment Analyzer}
\vspace{-0.1cm}
We develop a long short-term memory network (LSTM) based sentiment analyzer for crypto social media posts. LSTM enables the network to learn long-term relation, by utilizing forget and remember gates that allow the cell to decide which information to block or transmit based on its strength and importance. 

The social media post is first tokenized according to crypto word vocabulary and fed into the embedding layer, which converts the word token into the crypto word embedding. The LSTM based recurrent network is trained by taking the sequence of the embedding feature vector. A fully connected layer is used to transform the output of the LSTM later and activated with sigmoid to output the prediction. The labels of the posts used in training were manually labeled and encoded with positive (1), neutral (0) and negative (-1).  

\vspace{-0.2cm}
\section{Evaluation}
\vspace{-0.3cm}
We used the most recent 7 days' Sina-Weibo posts from top 100 crypto investors accounts as training data and the next 1 day's posts as testing. We use Precision and Recall to measure the performance of our LSTM sentiment predictor. Precision  measures the model's ability to return only relevant instances and recall measures the model's ability to classify all relevant instances.

We compare our method with the time series auto regression (AR) approach \cite{box2016} to evaluate the performance. As Table~\ref{tab:eval} shows, our approach outperforms the AR approach by 18.5\% in precision and 15.4\% in recall, exemplifies the effectiveness of the LSTM in analyzing the sentiment of social media content. 

\begin{table}[]
\vspace{-0.5cm}
\caption{Precision and recall on evaluating LSTM sentiment analyzer and AR.}
\centering
\begin{tabular}{|l|l|l|}
\hline
Method  & \textbf{Precision} & \textbf{Recall} \\
\hline
Auto Regression & 73.4\% & 80.2\% \\
\hline
LSTM Sentiment Analyzer & 87.0\%  & 92.5\% \\
\hline
\end{tabular}
\vspace{-0.4cm}
\label{tab:eval}
\end{table}

\vspace{-0.6cm}
\section{Conclusion}
\vspace{-0.2cm}
In this paper we introduce a crypto sentiment analyzer by utilizing the recurrent neural network to model the social media sentiment. The model is developed using LSTM and achieves higher precision and recall than the traditional auto regressive approach. The current sentiment analyzer can be used to predict the price fluctuation of the cryptocurrency and integrated to an autonomous trading system to assist the buying or selling of digital assets.
\vspace{-0.4cm}
\bibliographystyle{splncs04}
\bibliography{main}
%




\end{document}